\documentclass[twoside]{article}

\usepackage[accepted]{aistats2020}

\usepackage[utf8]{inputenc} % allow utf-8 input
\usepackage[T1]{fontenc}    % use 8-bit T1 fonts
\usepackage{hyperref}       % hyperlinks
\usepackage{url}            % simple URL typesetting
\usepackage{booktabs}       % professional-quality tables
\usepackage{amsfonts}       % blackboard math symbols
\usepackage{amssymb}
\usepackage{nicefrac}       % compact symbols for 1/2, etc.
\usepackage{microtype}      % microtypography
\usepackage{xcolor}
\usepackage{subcaption}
\usepackage{enumitem}
\usepackage{comment}
\usepackage{amsmath, amsfonts, amsthm, mathtools}
\usepackage{balance}
\usepackage[ruled,vlined]{algorithm2e}
\DeclarePairedDelimiter{\norm}{\lVert}{\rVert}
\DeclarePairedDelimiter{\inner}{\langle}{\rangle}

\newtheorem{theorem}{Theorem}[section]
\newtheorem{definition}{Definition}[section]
\newtheorem{lemma}{Lemma}[section]
\newtheorem{corollary}{Corollary}[section]

\begin{document}

% If your paper is accepted and the title of your paper is very long,
% the style will print as headings an error message. Use the following
% command to supply a shorter title of your paper so that it can be
% used as headings.
%
%\runningtitle{I use this title instead because the last one was very long}

% If your paper is accepted and the number of authors is large, the
% style will print as headings an error message. Use the following
% command to supply a shorter version of the authors names so that
% they can be used as headings (for example, use only the surnames)
%
\runningauthor{Chloe Ching-Yun Hsu, Michaela Hardt, Moritz Hardt}

\twocolumn[

\aistatstitle{Linear Dynamics: Clustering without identification}

\aistatsauthor{%
    Chloe Ching-Yun Hsu\textsuperscript{$\dagger$}\\
  %\texttt{chloehsu@berkeley.edu}
  \And
  Michaela Hardt\textsuperscript{$\dagger$}\\
  %\texttt{milaha@amazon.com}
  \And
  Moritz Hardt\textsuperscript{$\dagger$}\\
  %\texttt{hardt@berkeley.edu}
}

\aistatsaddress{ University of California, Berkeley \And Amazon \And University of California, Berkeley }
]

%\maketitle

\begin{abstract}
Linear dynamical systems are a fundamental and powerful parametric model class. However, identifying the parameters of a linear dynamical system is a venerable task, permitting provably efficient solutions only in special cases. This work shows that the eigenspectrum of unknown linear dynamics can be identified without full system identification.
We analyze a computationally efficient and provably convergent algorithm to estimate the eigenvalues of the state-transition matrix in a linear dynamical system.
    
When applied to time series clustering, our algorithm can efficiently cluster multi-dimensional time series with temporal offsets and varying lengths, under the assumption that the time series are generated from linear dynamical systems.
Evaluating our algorithm on both synthetic data and real electrocardiogram (ECG) signals, we see improvements in clustering quality over existing baselines.

\end{abstract}

\section{Introduction}\label{sec:intro}

Linear dynamical system (LDS) is a simple yet general model for time series. Many machine learning models are special cases of linear dynamical systems~\cite{roweis1999unifying}, including principal component analysis (PCA), mixtures of Gaussians, Kalman filter models, and hidden Markov models. 

When the states are hidden, LDS parameter identification has provably efficient solutions only in special cases, see for example~\cite{hazan2018spectral, hardt2018gradient, simchowitz2018learning}. In practice, the expectation–maximization (EM) algorithm~\cite{ghahramani1996parameter} is often used for LDS parameter estimation, but it is inherently non-convex and can often get stuck in local minima~\cite{hazan2018spectral}. 
Even when full system identification is hard, \textit{is there still hope to learn meaningful information about linear dynamics without learning all system parameters}? We provide a positive answer to this question. 
%Even when identification is hard, \textit{is there still hope to find meaningful clusters of linear systems without fully learning the systems}? We provide a positive answer to this question. 

We show that the eigenspectrum of the state-transition matrix of unknown linear dynamics can be identified without full system identification. The eigenvalues of the state-transition matrix play a significant role in determining the properties of a linear system. For example, in two dimensions, the eigenvalues determine the stability of a linear dynamical system. Based on the trace and the determinant of the state-transition matrix, we can classify a linear system as a stable node, a stable spiral, a saddle, an unstable node, or an unstable spiral.

To estimate the eigenvalues, we utilize a fundamental correspondence between linear systems and Autoregressive-Moving-Average (ARMA) models. We establish bi-directional perturbation bounds to prove that two LDSs have similar eigenvalues if and only if their output time series have similar auto-regressive parameters. Based on a consistent estimator for the autoregressive model parameters of ARMA models~\cite{tsay1984consistent}, we propose a regularized iterated least-squares regression method to estimate the LDS eigenvalues. Our method runs in time linear in the sequence length $T$ and converges to true eigenvalues at the rate $O_p(T^{-1/2})$.

As one application, our eigenspectrum estimation algorithm gives rise to a simple approach for time series clustering: First use regularized iterated least-squares regression to fit the autoregressive parameters; then cluster the fitted autoregressive parameters.

This simple and efficient clustering approach captures similarity in eigenspectrums, assuming each time series comes from an underlying linear dynamical system.
It is a suitable similarity measure where the main goal for clustering is to characterize state-transition dynamics regardless of change of basis, particularly relevant when there are multiple data sources with different measurement procedures.
Our approach bypasses the challenge of LDS full parameter estimation, while enjoying the natural flexibility to handle multi-dimensional time series with time offsets and partial sequences.

%Clustering is a useful tool to discover patterns from unlabeled time-series data, common in many domains such as sensor data from home, hospitals, particle accelerators, and oceans.
%However, clustering time series is a challenging task. Standard measures of similarity, such as the Euclidean distance, commonly used for clustering static data, fail to account for shifts and variable lengths of time series. Clustering instead the learned parameters of a dynamic model naturally overcomes these limitations, but often the underlying models can be difficult to learn from the time series observations. 

To verify that our method efficiently learns system eigenvalues on synthetic and real ECG data, we compare our approach to existing baselines,
including model-based approaches based on LDS, AR and ARMA parameter estimation, and PCA,
as well as model-agnostic clustering approaches such as dynamic time warping~\cite{cuturi2017soft} and k-Shape~\cite{paparrizos2015k}.

\textbf{Organization.}
We review LDS and ARMA models in Sec.~\ref{sec:prelim}.
In Sec.~\ref{sec:learning_eig} we discuss the main technical results around the correspondence between LDS and ARMA.
%In Sec.~\ref{sec:learning_eig} we show that outputs from $n$-dimensional LDS can be equivalently generated by an ARMA($n,n$) model, and that the autoregressive parameters from the ARMA($n,n$) model can provably recover the LDS eigenvalues.
Sec.~\ref{sec:armax_estimation} presents the regularized iterated regression algorithm, a consistent estimator of autoregressive parameters in ARMA models with applications to clustering.
We carry out eigenvalue estimation and clustering experiments on synthetic data and real ECG data in Sec.~\ref{sec:simulations}.
In the appendix, we describe generalizations to observable inputs and multidimensional outputs, and include additional simulation results.

\section{Related Work}

\textbf{Linear dynamical system identification.}
The LDS identification problem has been studied since the 60s~\cite{kalman1960new}, yet the theoretical bounds are still not fully understood. Recent provably efficient algorithms~\cite{simchowitz2018learning,hazan2018spectral,hardt2018gradient,dean2017sample} require setups that are not best-suited for time series clustering, such as assuming observable states and focusing on predition error instead of parameter recovery.
%Hazan et al.~\cite{hazan2018spectral, hazan2017learning}  proposed spectral filtering algorithms that minimize prediction error and in general do not identify system parameters. Hardt et al.~\cite{hardt2018gradient} show that gradient descent can identify system parameters under a strong assumption on the roots of the system. Simchowitz et al.~\cite{simchowitz2018learning} and Dean et al.~\cite{dean2017sample} proved new bounds for the LDS identification problem, with the assumption of observable states.
%We focus on LDSs with non-observable hidden states.

On recovering system parameters without observed states, Tsiamis et al. recently study a subspace identification algorithm with non-asymptotic $O(T^{-1/2})$ rate~\cite{tsiamis2019finite}. While our analysis does not provide finite sample complexity bounds, our simple algorithm achieves the same rate asymptotically.

\textbf{Model-based time series clustering.}
%Time-series clustering has been extensively studied in many areas including biology, climatology, energy consumption, finance, medicine, and voice recognition.
%Most existing time series clustering techniques fall into three approaches~\cite{liao2005clustering}: raw-data-based, feature-based, or model-based.
%Most existing techniques use one of the three major approaches: raw-data-based (e.g.~\cite{piciarelli2005trajectory, buzan2004extraction, morse2007efficient, chen2004marriage, chen2005robust,vlachos2002discovering, banerjee2001clickstream, kovsmelj1990cross, golay1998new, van1999cluster, kumar2002clustering, moller2003fuzzy, sakoe1990dynamic}), feature-based (e.g.~\cite{shaw1992using, goutte1999clustering, fu2001pattern, wilpon1985modified, vlachos2003wavelet} ), or model-based (e.g.~\cite{kalpakis2001distance, xiong2002mixtures, assfalg2006similarity, smyth1997clustering, li2011time, panuccio2002hidden, jansen1981piecewise, chen2007spade, biernacki2000assessing, maharaj2000cluster, li1999temporal, piccolo1990distance}).
%Raw-data based approaches have the downside of working directly with noisy high dimensional data, while feature-based approaches require domain-specific feature extraction. Model-based approaches have the drawback that underlying models might be hard to learn.
%Our approach is model-based, with linear dynamical system as our model.
Common model choices for clustering include Gaussian mixture models~\cite{biernacki2000assessing}, autoregressive integrated moving average (ARIMA) models~\cite{kalpakis2001distance}, and hidden Markov models~\cite{smyth1997clustering}. Gaussian mixture models, ARIMA models, and hidden Markov models are all special cases of the more general linear dynamical system model~\cite{roweis1999unifying}.
%, our work is a general model-based approach for time series clustering.

%A long line of work in control systems~\cite{hanzon1989identifiability} has studied the distance notion between LDSs, but most approaches are computationally expensive.
Linear dynamical systems have used to cluster video trajectories~\cite{chan2005probabilistic,afsari2012group,vishwanathan2007binet}, where the observed time series are higher dimensional than the hidden state dimension.
Our work is motivated by the more challenging sitatuion with a single or a few output dimensions, common in climatology, energy consumption, finance, medicine, etc.
%, with LDS similarity measures defined with the Kullback-Leibler (KL) divergence~\cite{chan2005probabilistic}, Binet-Cauchy Kernels~\cite{vishwanathan2007binet}, cepstra~\cite{de2000subspace, martin2000metric} and group theory~\cite{afsari2012group}.
%Afsari et al.~\cite{afsari2012group} defined a distance on the non-Euclidean space of LDSs, and demonstrated that the non-Euclidean LDS distance yields better clustering results than the Euclidean distance in computer vision applications. 

Compared to ARMA-parameter based clustering, our method only uses the AR half of the parameters which we show to enjoy more reliable convergence. We also differ from AR-model based clustering because fitting AR to an ARMA process results in biased estimates.

\textbf{Autoregressive parameter estimation.}
Existing spectral analysis methods for estimating AR parameters in ARMA models include high-order Yule-Walker (HOYW), MUSIC, and ESPRIT~\cite{stoica2005spectral,stoica1988high}. Our method is based on iterated regression~\cite{tsay1984consistent}, a more flexible method for handling observed exogenous inputs (see Appendix C) in the ARMAX generalization.
%This method is related to the Two–Stage Least Squares ARMA Method~\cite{mayne1982linear,stoica2005spectral}, while a major difference is that the iterated regression method is a consistent estimator for a fixed degree $p$ of the AR part, whereas the Two–Stage Least Squares ARMA method is only consistent as $p \rightarrow \infty$ (i.e. the asymptotic bias tends to 0 as $p \rightarrow \infty$).
%To the best of our knowledge, this is the first work to propose time-series clustering based on estimating only the AR parameters in ARMA.

\section{Preliminaries}\label{sec:prelim}
%In this section we review LDS and ARMA and state our assumptions. 

\subsection{Linear dynamical systems}
A discrete-time linear dynamical system (LDS) with parameters $\Theta = (A, B, C, D)$ receives inputs $x_1, \cdots, x_T \in \mathbb{R}^k$, has hidden states $h_0, \cdots, h_T \in \mathbb{R}^n$, and generates outputs $y_1, \cdots, y_T \in \mathbb{R}^m$ according to the following time-invariant recursive equations:
\begin{equation}\label{eq:lds_def}
\begin{aligned}
h_t = Ah_{t-1} + Bx_t + \zeta_t \\
y_t = Ch_t + Dx_t + \xi_t.
\end{aligned}
\end{equation}
\textbf{Assumptions.} 
We assume that the stochastic noise $\zeta_t$ and $\xi_t$ are diagonal Gaussians.
We also assume the system is observable, i.e. $C, CA, CA^2, \cdots, CA^{n-1}$ are linearly independent.
When the LDS is not observable, the ARMA model for the output series can be reduced to lower AR order, and there is not enough information in the output series to recover all the full eigenspectrum.

The model equivalence theorem (Theorem \ref{thm:model_eq}) and the approximation theorem (Theorem \ref{thm:approx_eig}) do not require any additional assumptions for any real matrix $A$. When additionally assuming $A$ only has simple eigenvalues in $\mathbb{C}$, i.e. each eigenvalue has multiplicity 1, we give a better convergence bound.

\textbf{Distance between linear dynamical systems.}
With the main goal to characterize state-transition dynamics, we view systems as equivalent up to change of basis, and use the $\ell_2$ distance of the spectrum of the transition matrix $A$, i.e. $d(\Theta_1, \Theta_2) = \norm{\lambda(A_1) - \lambda(A_2)}_2$, where $\lambda(A_1)$ and $\lambda(A_2)$ are the spectrum of $A_1$ and $A_2$ in sorted order. This distance definition satisfies non-negativity, identity, symmetry, and triangle inequality. 

Two very different time series could still have small distance in eigenspectrum. This is by design to allow the flexiblity for different measurement procedures, which mathematically correspond to different $C$ matricies.
When there are multiple data sources with different measurement procedures, our approach can compare the underlying dynamics of time series across sources.
%For example, ECG time series are measured between different electrodes, and climate data have temperatures, precipitation, along with other weather observations. While differently measured time series can look very different, their underlying state-transition matrices are still comparable.
%This eigenspectrum-based distance definition attempts to be agnostic of the specific measurement procedures, and admittedly as a tradeoff can have overly broad equivalence classes for certain applications.

\textbf{Jordan canonical basis.}
Every square real matrix is similar to a complex block diagonal matrix known as its Jordan canonical form (JCF). In the special case for diagonalizable matrices, JCF is the same as the diagonal form. Based on JCF, there exists a canonical basis $\{e_i\}$ consisting only of eigenvectors and generalized eigenvectors of $A$. A vector $v$ is a generalized eigenvector of rank $\mu$ with corresponding eigenvalue $\lambda$ if $(\lambda I - A)^{\mu}v = 0$ and $(\lambda I - A)^{\mu - 1}v \neq 0$.

\subsection{Autoregressive-moving-average models}

The autoregressive-moving-average (ARMA) model combines the autoregressive (AR) model and the moving-average (MA) model. The AR part involves regressing the variable with respect its lagged past values, while the MA part involves regressing the variable against past error terms.

\textbf{Autoregressive model.}
The AR model describes how the current value in the time series depends on the lagged past values. For example, if the GDP realization is high this quarter, the GDP in the next few quarters are likely high as well. An autoregressive model of order $p$, noted as AR$(p)$, depends on the past $p$ steps,
\begin{equation*}
y_t = c + \Sigma_{i=1}^p \varphi_i y_{t-i} + \epsilon_t,
\end{equation*}
where $\varphi_1, \cdots, \varphi_p$ are autoregressive parameters, $c$ is a constant, and $\epsilon_t$ is white noise.

When the errors are normally distributed, the ordinary least squares (OLS) regression is a conditional maximum likelihood estimator for AR models yielding optimal estimates~\cite{durbin1960estimation}. %We could also estimate AR parameters based on autocorrelation coefficients and Yule-Walker equations.

\textbf{Moving-average model.}
The MA model, on the other hand, captures the delayed effects of unobserved random shocks in the past. For example, changes in winter weather could have a delayed effect on food harvest in the next fall. A moving-average model of order $q$, noted as MA$(q)$, depends on unobserved lagged errors in the past $q$ steps,
\begin{equation*}
y_t = c + \epsilon_t + \Sigma_{i=1}^q \theta_i \epsilon_{t-i},
\end{equation*}
where $\theta_1, \cdots, \theta_q$ are moving-average parameters, $c$ is a constant, and the errors $\epsilon_t$ are white noise.

\textbf{ARMA model.}
The autoregressive-moving-average (ARMA) model, denoted as ARMA$(p,q)$, merges AR$(p)$ and MA$(q)$ models to consider dependencies both on past time series values and past unpredictable shocks,
\begin{equation*}
y_t = c + \epsilon_t + \Sigma_{i=1}^p \varphi_i y_{t-i} + \Sigma_{i=1}^q \theta_i \epsilon_{t-i}.
\end{equation*}

\textbf{ARMAX model.}
ARMA can be generalized to autoregressive–moving-average model with exogenous inputs (ARMAX).
\begin{equation*}
y_t = c + \epsilon_t + \Sigma_{i=1}^p \varphi_i y_{t-i} + \Sigma_{i=1}^q \theta_i \epsilon_{t-i} + \Sigma_{i=0}^r\gamma_ix_{t-i},
\end{equation*}
where $\{x_t\}$ is a known external time series, possibly multidimensional. In case  $x_t$ is a vector, the parameters $\gamma_i$ are also vectors.

Estimating ARMA and ARMAX models is significantly harder than AR, since the model depends on unobserved variables and the maximum likelihood equations are intractable~\cite{durbin1959efficient, choi2012arma}. Maximum likelihood estimation (MLE) methods are commonly used for fitting ARMA and ARMAX~\cite{guo1996self, bercu1995weighted, hannan1980estimation}, but have converge issues. Although regression methods are also used in practice, OLS is a \textit{biased} estimator for ARMA models~\cite{tiao1983consistency}. 

\textbf{Lag operator.}
We also introduce the lag operator, a concise way to describe ARMA models~\cite{granger1976time}, defined as $L y_t = y_{t-1}$. The lag operator could be raise to powers, or form polynomials. For example, $L^3 y_t = y_{t-3}$, and $(a_2 L^2 + a_1 L + a_0)y_t = a_2y_{t-2} + a_1y_{t-1} + a_0y_t$. The lag polynomials can be multiplied or inverted.
An AR($p$) model can be characterized by
\begin{equation*}
\Phi(L) y_t = c + \epsilon_t,
\end{equation*}
where $\Phi(L) = 1-\varphi_1 L - \cdots - \varphi_p L^p$ is a polynomial of the lag operator $L$ of degree $p$. For example, any AR(2) model can be described as $(1-\varphi_1 L -\varphi_2 L^2)y_t = c + \epsilon_t$.

Similarly, an MA($q$) can be characterized by a polynomial $\Psi(L) = \theta_qL^q + \cdots + \theta_1L + 1$ of degree $q$,
\begin{equation*}
y_t = c + \Psi(L)\epsilon_t.
\end{equation*}
For example, for an MA(2) model the equation would be $y_t = c + (\theta_2L^2 + \theta_1L + 1)\epsilon_t$.

Merging the two and adding dependency to exogenous input, we can write an ARMAX($p,q,r$) model as 
\begin{equation}\label{eq:arma_lag_op}
\Phi(L)y_t = c + \Psi(L)\epsilon_t + \Gamma(L)x_t
\end{equation}
where $\Phi, \Psi$, and $\Gamma$ are polynomials of degree $p, q$ and $r$. When the exogenous time series $x_t$ is multidimensional, $\Gamma(L)$ is a vector of degree-$r$ polynomials.

\section{Learning eigenvalues without system identification}\label{sec:learning_eig}

This section provides theoretical foundations for learning LDS eigenvalues from autoregressive parameters without full system identification. 
While general model equivalence between LDS and ARMA(X) is known~\cite{aastrom2013computer,kailath1980linear}, we provide detailed analysis of the exact correspondence between the LDS characteristic polynomial and the ARMA(X) autoregressive parameters along with perturbation bounds.

\subsection{Model equivalence}
We show that the output series from any LDS can be seen as generated by an ARMAX model,
whose AR parameters contain full information about the LDS eigenvalues.

\begin{theorem}\label{thm:model_eq}
Let $y_t \in \mathbb{R}^m$ be the outputs from a linear dynamical system with parameters $\Theta = (A,B,C,D)$, hidden dimension $n$, and inputs $x_t \in \mathbb{R}^k$. Each dimension of $y_t$ can be generated by an ARMAX($n, n, n-1$) model, whose autoregressive parameters $\varphi_1, \cdots, \varphi_n$ can recover the characteristic polynomial of $A$ by $\chi_A(\lambda) = \lambda^n - \varphi_1 \lambda^{n-1} - \cdots - \varphi_n$. 

In the special case where the LDS has no external inputs, the ARMAX model is an ARMA$(n,n)$ model.
\end{theorem}

See Appendix \ref{sec:model_eq_proof} for the full proof.

As a high-level proof sketch: We first analyze the hidden state projected to (generalized) eigenvector directions in Lemma~\ref{lemma:ar_poly_single}. We show that for a (generalized) eigenvector $e$ of the adjoint $A^*$ of the transition matrix with eigenvalue $\lambda$ and rank $\mu$, the time series obtained from applying the lag operator polynomial $(1-\lambda L)^{\mu}$ to $\inner{h_t, e}$ can be expressed as a linear combination of the past $k$ inputs $x_t, \cdots, x_{t-k+1}$. Since $A$ is real-valued, $A$ and its adjoint $A^*$ share the same characteristic polynomial $\chi_A$.

We then consider the lag operator polynomial $\chi_A^{\dagger}(L) = L^n\chi_A(L^{-1})$, and show that the time series obtained from applying $\chi_A^{\dagger}(L)$ applied to any (generalized) eigenvector direction is a linear combination of the past $k$ inputs. We use this on the Jordan canonical basis for $A^*$ that consists of (generalized) eigenvectors. From there, we conclude $\chi_A^{\dagger}$ is the autoregressive lag polynomial that contains the autoregressive coefficients.

The converse of Theorem \ref{thm:model_eq} also holds. An ARMA($p,q$) model can be seen as a $(p+q)$-dimensional LDS where the state encodes the relevant past values and error terms.
%, $h_t = (y_t, \cdots, y_{t-p+1}, \epsilon_t, \cdots, \epsilon_{t-q+1})$.

%An immediate corollary of Theorem \ref{thm:model_eq} is that the autoregressive parameters of system outputs contain full information about all non-zero eigenvalues of the system.

\begin{corollary}\label{cor:char_poly_determines_ar_params}
The output series of two linear dynamical systems have the same autoregressive parameters if and only if they have the same non-zero eigenvalues with the same multiplicities.
\end{corollary}

\begin{proof}
By Theorem \ref{thm:model_eq}, the autoregressive parameters are determined by the characteristic polynomial. Two LDSs of the same dimension have the same autoregressive parameters if and only if they have the same characteristic polynomials, and hence the same eigenvalues with the same multiplicities. Two LDSs of different dimensions $n_1 < n_2$ can have the same autoregressive parameters if and only if $\chi_{A_1}(\lambda) = \chi_{A_2}(\lambda) \lambda^{n_2-n_1}$ and $\varphi_{n_1+1} = \cdots = \varphi_{n_2} = 0$, in which case they have the same non-zero eigenvalues with same multiplicities.
\end{proof}

It is possible for two LDSs with different dimensions to have the same AR coefficients, if the higher-dimensional system has additional zero eigenvalues. Whether over-parameterized models indeed learn additional zero eigenvalues requires further empirical investigation.

\subsection{Approximation theorems for LDS eigenvalues}
We show that small error in the AR parameter estimation guarantees a small error in the eigenvalue estimation. This implies that an effective estimation algorithm for the AR parameters in ARMAX models leads to effective estimation of LDS eigenvalues.

\paragraph{General $(1/n)$-exponent bound}
\begin{theorem}\label{thm:approx_eig}
Let $y_t$ be the outputs from an $n$-dimensional linear dynamical system with parameters $\Theta=(A,B,C,D)$, eigenvalues $\lambda_1,\cdots,\lambda_n$, and hidden inputs. Let $\hat\Phi = (\hat\varphi_1, \cdots, \hat\varphi_n)$ be the estimated autoregressive parameters for $\{y_t\}$ with error $\norm{\hat\Phi - \Phi} = \epsilon$, and let $r_1, \cdots, r_n$ be the roots of the polynomial $1-\hat\varphi_1z-\cdots-\hat\varphi_nz^n$.

Assuming the LDS is observable, the roots converge to the true eigenvalues with convergence rate $\mathcal{O}(\epsilon^{1/n})$. 
If all eigenvalues of $A$ are simple (no multiplicity), then the convergence rate is $\mathcal{O}(\epsilon)$.
\end{theorem}

Without additional assumptions, the $\frac{1}{n}$-exponent in the above general bound is tight.
As an example, $z^2 - \epsilon$ has roots $z \pm \sqrt{\epsilon}$. The general phenomenon that a root with multiplicity $m$ can split into $m$ roots at rate $O(\epsilon^m)$ is related to the regular splitting property~\cite{hryniv1999perturbation, lancaster2003perturbation}.
%in matrix eigenvalue perturbation theory. 

\paragraph{Linear bound for simple eigenvalues}\mbox{}\\
\vspace{-1.5em}

Under the additional assumption that all the eigenvalues are simple (no multiplicity), we derive a better $O(\epsilon)$ bound instead of $O(\epsilon^{1/n})$.
We show small perturbation in AR parameters results in small perturbation in companion matrix, and small perturbation in companion matrix results in small perturbation in eigenvalues.

We defer the full proofs to Appendix~\ref{sec:approx_eig}, but describe the proof ideas here.

For a monic polynomial $\Phi(u) = z^n + \varphi_1 z^{n-1} + \cdots + \varphi_{n-1}z + \varphi_n$,
the \textit{companion matrix}, also known as the controllable canonical form in control theory, is the square matrix
$$C(\Phi)={\begin{bmatrix}0&0&\dots &0&-\varphi_{n}\\1&0&\dots &0&-\varphi_{n-1}\\0&1&\dots &0&-\varphi_{n-2}\\\vdots &\vdots &\ddots &\vdots &\vdots \\0&0&\dots &1&-\varphi_{1}\end{bmatrix}}.$$
The matrix $C(\Phi)$ is the companion in the sense that it has $\Phi$ as its characteristic polynomial.

%When $\Phi$ has distinct roots $\lambda_1, \cdots, \lambda_n$, the companion matrix $C(\Phi)$ is diagonalizable as
%\begin{equation}
%C(\Phi) = V^{-1}\text{diag}(\lambda_1, \cdots, \lambda_n)V,
%\end{equation}
%where $V$ is the Vandermonde matrix corresponding to the $\lambda$'s. Note that $V$ is not an orthogonal matrix.

In relation to an autoregressive AR($p$) model, the companion matrix corresponds to the transition matrix in the linear dynamical system when we encode the values form the past $p$ lags as a $p$-dimensional state
$$h_t = {\begin{bmatrix}y_{t-p+1} & \cdots & y_{t-1} & y_t\end{bmatrix}}^T.$$ If $y_{t} = \varphi_1 y_{t-1} + \cdots + \varphi_p y_{t-p}$, then $h_{t} =$
\begin{equation}\label{eq:sys_from_companion}
\begin{split}
&{\begin{bmatrix}y_{t-p+1}\\y_{t-p+2}\\ \cdots \\ y_{t-1} \\ y_{t}\end{bmatrix}}
= {\begin{bmatrix}0&1&0&\dots &0\\0&0&1&\dots &0\\ \vdots&\vdots&\vdots&\ddots&\vdots \\0&0&0&\dots &1\\ \varphi_p&\varphi_{p-1}&\varphi_{p-2}&\dots &\varphi_{1}\end{bmatrix}}
  {\begin{bmatrix}y_{t-p}\\y_{t-p+1}\\ \cdots \\ y_{t-2} \\ y_{t-1}\end{bmatrix}} \\
&= C(-\Phi)^T h_{t-1}.
\end{split}
\end{equation}

We then use matrix eigenvalue perturbation theory results on the companion matrix for the desired bound.

\begin{lemma}[Theorem 6 in~\cite{lancaster2003perturbation}]\label{lemma:semisimple_perturb}
Let $L(\lambda, \epsilon)$ be an analytic matrix function with semi-simple eigenvalue $\lambda_0$ at $\epsilon=0$ of multiplicity $M$. Then there are exactly $M$ eigenvalues $\lambda_i(\epsilon)$ of $L(\lambda, \epsilon)$ for which $\lambda_i(\epsilon) \rightarrow \lambda_0$ as $\epsilon \rightarrow 0$, and for these eigenvalues
\begin{equation}\label{eq:eig_split}
\lambda_{i}(\epsilon) = \lambda_0 + \lambda_{i}'\epsilon + o(\epsilon).
\end{equation}
\end{lemma}

\paragraph{Explicit bound on condition number}\mbox{}\\
\vspace{-1.5em}

When the LDS has all simple eigenvalues, we provide a more explicit bound on the condition number. 

\begin{theorem}\label{thm:explicit_cond_num}
In the same setting as above in Theorem \ref{thm:approx_eig}, when all eigenvalues of $A$ are simple, $|r_j - \lambda_j| \leq \kappa\epsilon + o(\epsilon^2)$,
% \begin{equation}
% |r_j - \lambda_j| \leq \kappa\epsilon + o(\epsilon^2),
% \end{equation}
then the condition number $\kappa$ is bounded by
\begin{equation*}\label{eq:cond_num_bound}
 \begin{aligned}
\frac{1}{\prod_{k \neq j} |\lambda_j - \lambda_k|} \leq
\kappa \leq \\
\frac{\sqrt{n}}{\prod_{k \neq j} |\lambda_j - \lambda_k|} \left(\max(1, |\lambda_j|)\right)^{n-1} (1 + \rho(A)^2)^{\frac{n-1}{2}},
\end{aligned}
\end{equation*}
where $\rho(A)$ is the spectral radius, i.e. largest absolute value of its eigenvalues.

In particular, when $\rho(C) \leq 1$, i.e. when the matrix is Lyapunov stable, then the absolute difference between the root from the auto-regressive method and the eigenvalue is bounded by
$
|r_j - \lambda_j| \leq \frac{\sqrt{n} (\sqrt{2})^{n-1}}{\prod_{k \neq j} |\lambda_j - \lambda_k|}\epsilon + o(\epsilon^2).
$
\end{theorem}
In Appendix~\ref{sec:approx_eig}, we derive the explicit formula in Theorem~\ref{thm:explicit_cond_num} by conjugating the companion matrix by a Vandermonde matrix to diagonalize it and invoking the explicit inverse formula of Vandermonde matrices.

\section{Estimation of ARMA autoregressive parameters}\label{sec:armax_estimation}

In general, learning ARMA models is hard, since the output series depends on unobserved error terms. Fortunately, for our purpose we are only interested in the autoregressive parameters, that are easier to learn since the past values of the time series are observed.

The autoregressive parameters in an ARMA($p,q$) model are not equivalent to the pure AR($p$) parameters for the same time series. For AR($p$) models, ordinary least squares (OLS) regression is a consistent estimator of the autoregressive parameters~\cite{lai1983asymptotic}. However, for ARMA($p, q$) models, due to the serial correlation in the error term $\epsilon_t + \sum_{i=1}^q \theta_i \epsilon_{t-i}$, the OLS estimates for autoregressive parameters can be biased~\cite{tiao1983consistency}.

\textbf{Regularized iterated regression.}
Iterated regression~\cite{tsay1984consistent} is a consistent estimator for the AR parameters in ARMA models. While iterated regression is theoretically well-grounded,
it tends to over-fit and results in excessively large parameters.
To avoid over-fitting, we propose a slight modification with regularization, which keeps the same theoretical guarantees and yields better practical performance.

We also generalize the method to handle \textit{multidimensional outputs} from the LDS and  \textit{observed inputs} by using ARMAX instead of ARMA models, as described in details in Appendix \ref{sec:appendix_armax_estimation} as Algorithm \ref{algo:iter_reg_general}. 

\begin{algorithm}
\SetAlgoLined
Input: Time series $\{y_t\}_{t=1}^{T}$, target hidden state dimension $n$, and regularization coefficient $\alpha$. \\
Initialize error term estimates $\hat\epsilon_t = 0$ for $t=1,\dots, T$\;
\For{$i=0,\cdots,n$} {
Perform $\ell_2$-regularized least squares regression to estimate $\hat\varphi_j$, $\hat\theta_j$, and $\hat c$ in
$y_t = \sum_{j=1}^n \hat\varphi_j y_{t-j} + \sum_{j=1}^i \hat \theta_j \hat\epsilon_{t-j} + \hat c$ with regularization strength $\alpha$ only on the $\hat\theta_j$ terms\;
Update $\hat\epsilon_t$ to be the residuals from the most recent regression\;
}
Return $\hat\varphi_1, \cdots, \hat\varphi_n$.
 \caption{Regularized iterated regression for autoregressive parameter estimation
}\label{algo:iter_reg_1d}
\end{algorithm}

The $i$-th iteration of the regression only uses error terms from the past $i$ lags. The initial iteration is an ARMA($n,0$) regression, the first iteration is an ARMA($n,1$) regression, and so forth until ARMA($n, n$) in the last iteration.

\textbf{Time complexity.}
The iterated regression involves $n+1$ steps of least squares regression each on at most $2n+1$ variables. Therefore, the total time complexity of Algorithm \ref{algo:iter_reg_1d} is $O(n^3T + n^4)$, where $T$ is the sequence length and $n$ is the hidden state dimension.

\textbf{Convergence rate.}
The consistency and the convergence rate of the estimator is analyzed in \cite{tsay1984consistent}.
Adding regularization does not change the asymptotic property of the estimator.
\begin{theorem}[\cite{tsay1984consistent}]\label{thm:armax_convergence}
Suppose that $y_t$ is an ARMA($p, q$) process, stationary or not. The estimated autoregressive parameters $\hat\Phi = (\hat\varphi_1, \cdots, \hat\varphi_n)$ from iterated regression converges in probability to the true parameters with rate
\begin{equation*}
\hat\Phi = \Phi + O_p(T^{-1/2}),
\end{equation*}
or more explicitly, convergence in probability means that for all $\epsilon$, $\lim_{T\rightarrow\inf} \Pr(T^{1/2} |\hat\Phi - \Phi| > \epsilon) = 0.$
\end{theorem}

%The main application discussed here is time-series clustering. Other than clustering, the regularized iterated regression method for estimating AR parameters in ARMA/ARMAX also has potential applications as a feature engineering step for supervised time series classification and prediction tasks, or as a preliminary estimation step for initialization in full LDS system identification.

\subsection{Applications to clustering}\label{sec:clustering}
The task of clustering depends on an appropriate distance measure for the clustering purpose.
When there are multiple data sources for time series with comparable dynamics but measured by different measurement procedures, one might hope to cluster time series based only on the state-transition dynamics.

We can observe from the LDS definition \eqref{eq:lds_def} that two LDSs with parameters $(A,B,C,D)$ and $(A',B',C',D')$ are equivalent if $A' = P^{-1}AP, B'=P^{-1}B, C'=CP, D'=DP$, and $h_t' = P^{-1}h_t$ under change of basis by some non-singular matrix $P$. Therefore, to capture the state-transition dynamics while allowing for flexibility in the measurement matrix $C$, the distance measure should be invariant under change of basis.
We choose to use the eigenspectrum distance between state-transition matrices, a natural distance choice that is invariant under change of basis.

%The assumption of a latent LDS is general since many classic models such as PCA, mixtures of Gaussians, and hidden Markov models are special cases of LDSs.

In previous sections, our theoretical analysis shows that AR parameters in ARMA time series models can effectively estimate the eigenspectrum of underlying LDSs.
We therefore propose a simple time series clustering algorithm: 1) first use iterated regression to estimate the autoregressive parameters in ARMA models for each times series, and 2) then apply any standard clustering algorithm such as K-means on the distance between autoregressive parameters.
%For LDS with high-dimensional hidden states, clustering high-order AR parameters might be more challenging for K-means, and one possible workaround is to cluster only based on the top-K eigenvalues.

Our method is very flexible. It handles multi-dimensional data, as Theorem~\ref{thm:model_eq} suggests that any output series from the same LDS should share the same autoregressive parameters. It can also handle exogenous inputs as illustrated in Algorithm~\ref{algo:iter_reg_general} in Appendix~\ref{sec:appendix_armax_estimation}. It is scale, shift, and offset invariant, as the autoregressive parameters in ARMA models are. It accommodates missing values in partial sequences as we can still perform OLS after dropping the rows with missing values. It also allows sequences to have different lengths, and could be adapted to handle sequences with different sampling frequencies, as the compound of multiple steps of LDS evolution is still linear.

\begin{table*}[!t]
\small
\centering
  \begin{tabular}{p{1.6cm}lllll}
\toprule
      \# Clusters  & Method &  Adj.~Mutual~Info.  &  Adj.~Rand~Score & V-measure & Runtime~(secs)\\
\midrule
2  & \texttt{AR} &    0.06 (0.04-0.08) &    0.07 (0.05-0.09) &  0.07 (0.05-0.09) &        1.09 (1.01-1.17) \\
   & \texttt{ARMA} & \textbf{0.13 (0.11-0.16)} & \textbf{0.16 (0.13-0.19)} & \textbf{0.14 (0.11-0.16)} &  \textbf{0.44 (0.40-0.47)} \\
   & \texttt{ARMA\_MLE} &  0.02 (0.01-0.03) &  0.02 (0.01-0.03) &  0.03 (0.02-0.04) &  70.64 (68.01-73.28) \\
   & \texttt{DTW} &    0.02 (0.01-0.03) &    0.02 (0.01-0.03) &  0.03 (0.02-0.04) &        6.60 (6.34-6.86) \\
   & \texttt{k-Shape} &    0.03 (0.02-0.04) &    0.03 (0.02-0.05) &  0.04 (0.03-0.05) &     28.58 (25.15-32.01) \\
   & \texttt{LDS} &    0.09 (0.06-0.12) &    0.09 (0.06-0.12) &  0.10 (0.07-0.12) &  341.08 (328.24-353.93) \\
   & \texttt{PCA} &  -0.00 (-0.00-0.00) &  -0.00 (-0.00-0.00) &  0.02 (0.02-0.02) &        0.45 (0.43-0.47) \\
3  & \texttt{AR} &    0.11 (0.09-0.12) &    0.09 (0.07-0.10) &  0.12 (0.11-0.14) &        1.02 (0.93-1.10) \\
   & \texttt{ARMA} &    0.18 (0.16-0.20) &    0.16 (0.14-0.18) &  0.19
      (0.17-0.21) & \textbf{0.42 (0.38-0.46)} \\
   & \texttt{ARMA\_MLE} &  0.04 (0.03-0.05) &  0.04 (0.03-0.05) &  0.06 (0.05-0.07) &  72.58 (69.63-75.52) \\
   & \texttt{DTW} &    0.04 (0.03-0.05) &    0.03 (0.02-0.03) &  0.07 (0.06-0.08) &        6.63 (6.37-6.90) \\
   & \texttt{k-Shape} &    0.06 (0.05-0.07) &    0.04 (0.04-0.05) &  0.08 (0.07-0.09) &     40.10 (34.65-45.55) \\
      & \texttt{LDS} & \textbf{0.20 (0.18-0.23)} &  \textbf{0.17 (0.15-0.20)} &
      \textbf{0.22 (0.19-0.24)} &  338.67 (325.66-351.67) \\
   & \texttt{PCA} &   0.00 (-0.00-0.00) &   0.00 (-0.00-0.00) &  0.04 (0.04-0.04) &        0.47 (0.45-0.49) \\
5  & \texttt{AR} &    0.17 (0.16-0.18) &    0.11 (0.10-0.12) &  0.22 (0.21-0.23) &        0.91 (0.83-1.00) \\
   & \texttt{ARMA} &    0.22 (0.21-0.23) &    0.15 (0.14-0.16) &  0.26
      (0.25-0.28) & \textbf{0.40 (0.35-0.45)} \\
   & \texttt{ARMA\_MLE} &  0.08 (0.07-0.09) &  0.05 (0.04-0.05) &  0.14 (0.13-0.15) &  74.00 (71.70-76.30) \\
   & \texttt{DTW} &    0.05 (0.04-0.06) &    0.03 (0.02-0.03) &  0.11 (0.10-0.12) &        6.20 (5.86-6.55) \\
   & \texttt{k-Shape} &    0.08 (0.07-0.09) &    0.05 (0.04-0.05) &  0.14 (0.13-0.15) &    97.83 (84.97-110.68) \\
      & \texttt{LDS} &  \textbf{0.25 (0.23-0.26)} &  \textbf{0.17 (0.16-0.18)} &
      \textbf{0.29 (0.28-0.30)} &  321.40 (304.17-338.64) \\
   & \texttt{PCA} &    0.01 (0.01-0.01) &    0.00 (0.00-0.01) &  0.08 (0.07-0.08) &        0.52 (0.49-0.55) \\
10 & \texttt{AR} &    0.22 (0.21-0.22) &    0.11 (0.11-0.12) &  0.38 (0.37-0.38) &        0.87 (0.77-0.98) \\
 & \texttt{ARMA} & \textbf{0.24 (0.23-0.25)} &    \textbf{0.14 (0.13-0.15)}
 &  \textbf{0.39 (0.39-0.40)} &    \textbf{0.42 (0.37-0.48)} \\
 & \texttt{ARMA\_MLE} &  0.11 (0.10-0.12) &  0.06 (0.05-0.06) &  0.29 (0.28-0.30) &  63.39 (60.49-66.30) \\
   & \texttt{DTW} &    0.06 (0.06-0.07) &    0.03 (0.02-0.03) &  0.25 (0.24-0.25) &        5.51 (5.03-5.98) \\
   & \texttt{k-Shape} &    0.08 (0.07-0.08) &    0.04 (0.03-0.04) &  0.26 (0.26-0.27) &   108.79 (93.41-124.18) \\
      & \texttt{LDS} &    0.23 (0.22-0.24) &    0.13 (0.12-0.13) &  \textbf{0.39
      (0.38-0.40)} &  277.45 (253.38-301.51) \\
   & \texttt{PCA} &    0.02 (0.01-0.02) &    0.01 (0.00-0.01) &  0.14 (0.13-0.15) &        0.47 (0.44-0.51) \\
\bottomrule
\end{tabular}
  \vspace{0.5em}
  \caption{Performance of clustering 100 random 2-dimensional LDSs based on their output series of length 1000, with 95\% confidence intervals from 100 trials. AMI, Adj. Rand, and V-measure are the adjusted mutual information score, adjusted Rand score, and V-measure between ground truth cluster labels and learned clusters. The runtime is on an instance with 12 CPUs and 40 GB memory running Ubuntu 18.}
  \label{table:2d_ami}
  \vspace{-1em}
\end{table*}

\begin{figure*}[!t]
  \centering
  %\fbox{\rule[-.5cm]{0cm}{4cm} \rule[-.5cm]{4cm}{0cm}}
  \includegraphics[width=0.65\linewidth]{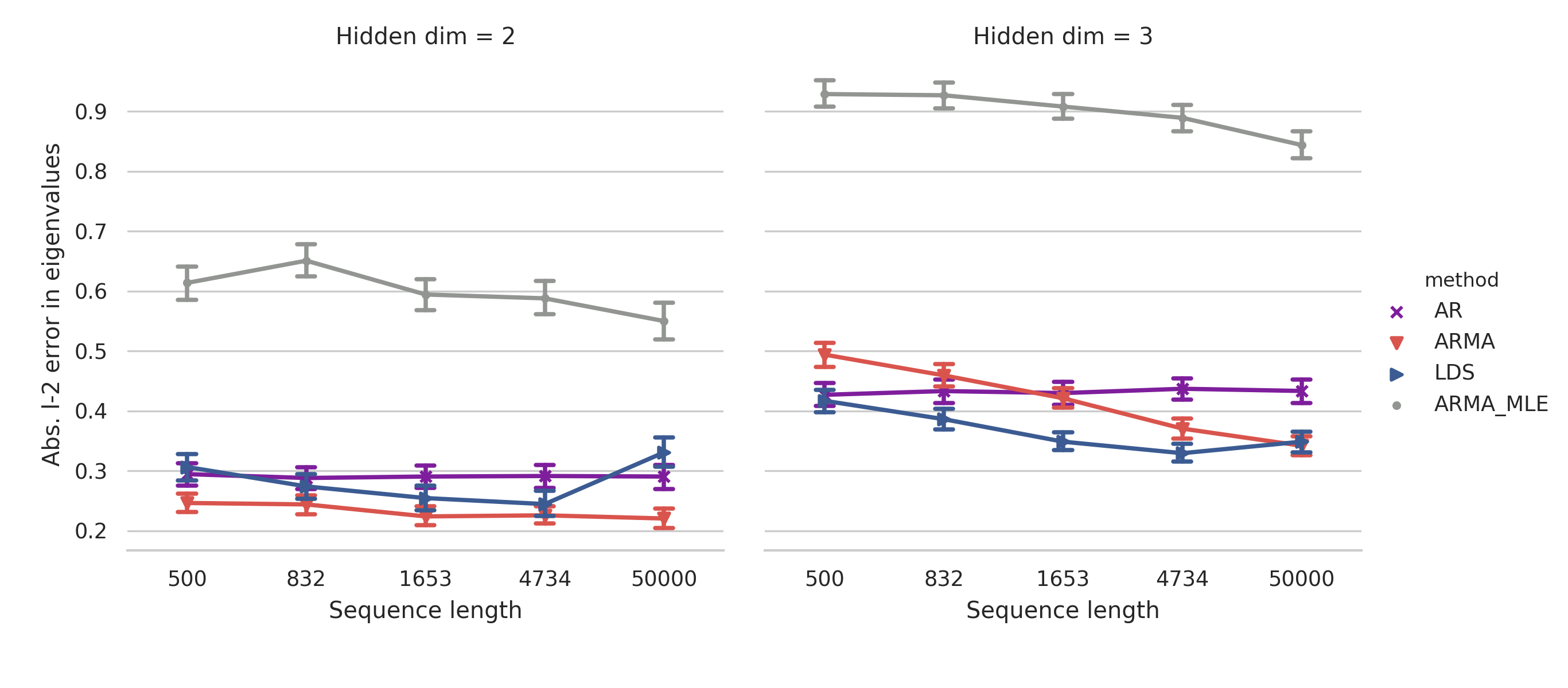}
  \vspace{-1em}
  \caption{Absolute $\ell_2$-error in eigenvalue estimation for 2-dimensional and 3-dimensional LDSs, with 95\%
    confidence interval from 500 trials.}\label{fig:learn_eig}
\end{figure*}

\begin{table*}[t!]
  \small
  \centering
  \begin{tabular}{lllll}
    \toprule
    Method     & AMI & Adj.~Rand Score & V-measure  \\
\midrule
\texttt{ARMA}  &  0.12 (0.10-0.13) &  0.12 (0.11-0.14) &  0.14 (0.12-0.16)  \\
% ARMA\_MLE  &  0.04 (0.03-0.05) &  0.03 (0.02-0.04) &  0.07 (0.05-0.08) &  21.63 (20.91-22.34) \\
\texttt{AR}        &  0.10 (0.09-0.12) &  0.10 (0.09-0.12) &  0.13 (0.11-0.14)  \\
\texttt{PCA}       &  0.04 (0.03-0.05) &  0.02 (0.02-0.03) &  0.07 (0.06-0.08)  \\
\texttt{LDS} &  0.09 (0.07-0.10) &  0.11 (0.10-0.13) &  0.10 (0.09-0.11)  \\
\texttt{k-Shape}    &  0.08 (0.07-0.10) &  0.09 (0.07-0.11) &  0.10 (0.09-0.12)  \\
\texttt{DTW}    &   0.03 (0.02-0.04) & 0.02 (0.01-0.03) & 0.06 (0.05-0.07) \\
%\texttt{GAK}    &   0.04 (0.03-0.05) & 0.04 (0.03-0.05) & 0.06 (0.05-0.07) \\
\bottomrule
  \end{tabular}
  \vspace{0.2em}
  \caption{Clustering performance on electrocardiogram (ECG) data separating segments of normal sinus rhythm from supraventricular tachycardia. 95\% Confidence intervals are from 100 bootstrapped samples of 50 series.} \label{table:ecg}
  \vspace{-1em}
\end{table*}

\section{Experiments}\label{sec:simulations}
We experimentally evaluate the quality and efficiency of the clustering from our method and compare it to existing baselines.
The source code is available online at
\url{https://github.com/chloechsu/ldseig}.
%\url{https://github.com/google-research/google-research/tree/master/linear_dynamical_systems}.

%We start out with simulated data generated from LDSs in Sec.~\ref{sec:sim} satisfying the assumptions of our method.
%Then we turn to real ECG data in Sec.~\ref{sec:real_experiments} to see if our method works in practice as a proof of concept. 

\subsection{Methods}\label{sec:methods}
%We compare the following approaches.

\begin{itemize}[noitemsep,topsep=0pt,parsep=0pt,partopsep=0pt]
  \item \texttt{ARMA:} K-means on AR parameters in ARMA($n,n$) model estimated by regularized iterated regression as we proposed in Algorithm~\ref{algo:iter_reg_1d}.
  \item \texttt{ARMA\_MLE:} K-means on AR parameters in ARMA($n,n$) model estimated by the MLE method using statsmodels~\cite{seabold2010statsmodels}.
  \item \texttt{AR:} K-means on AR parameters in AR($n$) model estimated by OLS using statsmodels.
  \item \texttt{LDS:} K-means on estimated LDS eigenvalues. We estimate the LDS eigenvalues with the pylds package~\cite{pylds}, with 100 EM steps initialized by 10 Gibbs iterations.
  \item \texttt{k-Shape:} A shape-based time series clustering method~\cite{paparrizos2015k}, using the tslearn package~\cite{tavenardtslearn}.
  \item \texttt{DTW}\texttt{:} K-medoids on dynamic time warping distance, using the dtaidistance~\cite{dtaidistance} and pyclustering~\cite{novikov2019pyclustering} packages.
  \item \texttt{\texttt{PCA}:} K-means on the first $n$ PCA components, using sklearn~\cite{pedregosa2011scikit}. 
  %\item \texttt{GAK:} K-means on Global Alignment Kernel~\cite{cuturi2011fast}, using tslearn.
\end{itemize}

\subsection{Metrics of Cluster Quality}
We measure cluster quality using three metrics in sklearn: V-measure~\cite{rosenberg2007v}, adjusted mutual information~\cite{vinh2010information}, and adjusted Rand score~\cite{hubert1985comparing}.
%All three metrics compare learned clusters with ground truth clusters.
%AMI adjusts the mutual information score to account for the number of clusters that tends to increase MI. The AMI metric is symmetric and independent of permutation of label values.

\subsection{Simulation}\label{sec:sim}
%We generate data following the assumptions behind our method. We study clustering quality and efficiency across methods and provide a deeper dive into the underlying eigenvalue estimation. 

\textbf{Dataset.}
We generate LDSs representing cluster centers with random matrices of i.i.d. Gaussian entries.
From the cluster centers, we derive LDSs that are close to the cluster centers. From each LDS, we generate time series of length 1000 by drawing inputs from standard Gaussians and adding noise to the output sampled from $N(0, 0.01^2)$. More details in Appendix~\ref{sec:data_generation}.

\textbf{Clustering performance.}
The iterated \texttt{ARMA} regression method and the \texttt{LDS} method yield the best clustering quality, while the iterated \texttt{ARMA} regression method is significantly faster.
These results hold up for choices of different cluster quality metrics and number of clusters.

% \subsubsection{Determining the number of clusters}
% In the above clustering results, we assume knowledge of the true number of clusters. When the number of clusters is unknown, we could combine the well-known Silhouette method with autoregressive parameter estimation for determining the number of clusters by optimizing the Silhouette score. The Sihouette score is a clustering quality metric based on distance, and does not require knowledge of ground truth cluster labels. After learning the autoregressive parameters, we compute the average Sihouette score based on the distance between autoregressive parameters, and pick the number of clusters accordingly.

% As an example, we generate 3 well-separated clusters of 150 total LDSs where the cluster centers have eigenvalues $(0.5, 0.3)$, $(-0.5, -0.3)$, and $(0.9, 0.8)$. The 

\textbf{Eigenvalue Estimation.}
Good clustering results rely on good approximations of the LDS eigenvalue distance. 
Our analyses in Theorem \ref{thm:armax_convergence} and Theorem \ref{thm:approx_eig} proved that the iterative \texttt{ARMA} regression algorithm can learn the LDS eigenvalues with converge rate $O_p(T^{-1/2})$. In~Figure \ref{fig:learn_eig},  we see that the observed convergence rate in simulations roughly matches the theoretical bound.

Each EM step in \texttt{LDS} runs in $O(n^3T)$. When running a constant number of EM steps, \texttt{LDS} has the same total complexity as iterated \texttt{ARMA}.
We chose 100 steps based on empirical evaluation of convergence for sequence length 1000.
However, longer sequences may need more EM steps to converge, which would explain the increase in \texttt{LDS} eigenvalue estimation error for sequence length 50000 in Figure~\ref{fig:learn_eig}.
Depending on different implementations and initialization schemes, it is possible that the \texttt{LDS} performance can be further optimized. 

\texttt{ARMA} and \texttt{LDS} have comparable eigenvalue estimation error for most configurations.
While the pure \texttt{AR} approach also gives comparable estimation error on relatively short sequences, its estimation is biased, and the error does not go down as sequence length increases.

% For longer sequences, iterative ARMA regression outperforms pure AR regression. The Gibbs sampling method for LDS MLE estimation has comparable performance with iterative ARMA regression, but it is 2 orders of magnitude slower (see Figure \ref{fig:learn_eig_runtime} in Appendix \ref{sec:additional_sumulations}). The ARMA MLE estimation method results in much higher error, and hence is omitted from the plot.

% \textbf{Overparametrization.}
% %\subsection{Time-series data with mixed sequence length}
% %\subsection{Time-series data with mixed hidden dimensions}

% \textbf{Deep dive case studies.}
% Discussion about LDS spectrum; visualization of intra/inter-cluster distances and other visualization plots on 2D examples.

\subsection{Real-world ECG data}\label{sec:real_experiments}
While our simulation results show the efficacy of our method, the data generation process satisfy assumptions that may not hold on real data. 
As a proof-of-concept, we also test our method on real electrocardiogram (ECG) data.

\textbf{Dataset.}
The MIT-BIH~\cite{mit-bih} dataset in PhysioNet~\cite{PhysioNet} is the most common dataset for evaluating algorithms for ECG data~\cite{de2004automatic,yeh2012analyzing,ozbay2006fuzzy,ceylan2009novel}.

It contains 48 half-hour recordings collected at the Beth Israel Hospital between 1975 and 1979. 
%We use the MIT-BIH dataset from physionet~\cite{mit-bih,PhysioNet}, the most common dataset used to design and evaluate ECG  algorithms \cite{yeh2012analyzing,ozbay2006fuzzy,ceylan2009novel,syed2007clustering,korurek2010clustering,de2004automatic}. It contains 48 half-hour recordings collected at the Beth Israel Hospital between 1975 and 1979. 
Each two-channel recording is digitized at a rate of 360 samples per second per channel. 15 distinct rhythms are annotated in recordings including abnormalities of cardiac rhythm (arrhythmias) by two cardiologists. 

Detecting cardiac arrhythmias has stimulated research and product applications such as Apple's FDA-approved detection of atrial fibrillation~\cite{apple-ecg}.
ECG data have been modeled with AR and ARIMA models~\cite{ARIMA-ECG,Corduas2008TimeSC,AR-ECG}, and more recently convolutional neural networks~\cite{arrhythmia-cnn}.

We bootstrap 100 samples of 50 time series; each bootstrapped sample consists of 2 labeled clusters: 25 series with supraventricular tachycardia and 25 series with normal sinus rhythm. Each series has length 500 which adequately captures a complete cardiac cycle. We set the \texttt{ARMA} $\ell_2$-regularization coefficient to be 0.01, chosen based on our simulation results.
%For \texttt{DTW} and \texttt{GAK}, further subsampling improves performance so we report best metrics from subsampling to sequence length 100.

\textbf{Results.}
Comparing methods outlined in Section~\ref{sec:methods}, Table~\ref{table:ecg} shows that our method achieves the best quality closely followed by the \texttt{AR} and \texttt{LDS} methods, according to adjusted mutual information, adjusted Rand score and V-measure, while being computationally efficient.

%To put our real-world experimental results on ECG data into perspective, the adjusted rand score of 0.12 is not too far off from the best score of 0.16 observed in simulation experiments even when LDS is the ground truth (Table 3-4 in Appendix D). Clustering time series based on their underlying dynamics is an inherently challenging task due to the unobserved latent states. 

%The main improvement of \texttt{ARMA} over \texttt{AR} is to correct for the bias in \texttt{AR} due to serial correlation in error terms. In our ECG experiment the bias effect is small, but may increase further for longer sequences.
%See Figure 1 and Figure 2 (Appendix~\ref{sec:additional_sumulations}) for simulation results on varying sequence lengths.
%While the \texttt{LDS} method is only slightly worse than \texttt{ARMA}, it is 200+ times slower, and more complicated to implement. 

% \section{Discussion}
% \subsection{Generalization to time-series data with observed inputs}

\section{Conclusion}
We give a fast, simple, and provably effective method to estimate linear dynamical system (LDS) eigenvalues based on system outputs.
The algorithm combines statistical techniques from the 80's with our insights on the correspondence between LDSs and ARMA models.

As a proof-of-concept, we apply the eigenvalue estimation algorithm to time series clustering.
The resulting clustering approach is flexible to handle varying lengths, temporal offsets, as well as multidimensional inputs and outputs.
%Specifically, we show that two LDSs have similar eigenvalues if and only if their generated time series have similar auto-regressive parameters in ARMA models.
Our efficient algorithm yields high quality clusters in simulations and on real ECG data. 

While LDSs are general models encompassing mixtures of Gaussian and hidden Markov models, they may not fit all applications. It would be interesting to extend the analysis to non-linear models, and to consider model overparameterization and misspecification.

\paragraph{Acknowledgments}
We thank the anonymous reviewers for thoughtful comments, and Scott Linderman, Andrew Dai, and Eamonn Keogh for helpful guidance.
%We thank the anonymous reviewers for their thoughtful comments, especially for pointing us to relevant related work and suggesting further directions to explore. We also would like to thank Scott Linderman and Eamonn Keogh for helpful guidance on PyLDS and DTW, and Andrew Dai for early feedback on the manuscript.

\balance
\bibliography{citations}
\newpage
\normalsize

\appendix

\section{Proofs for model equivalence}\label{sec:model_eq_proof}

In this section, we prove a generalization of Theorem \ref{thm:model_eq} for both LDSs with observed inputs and LDSs with hidden inputs.

\subsection{Preliminaries}\label{sec:appendix_model_eq_prelim}

\paragraph{Sum of ARMA processes}
It is known that the sum of ARMA processes is still an ARMA process.

\begin{lemma}[Main Theorem in~\cite{granger1976time}]\label{lemma:sum_arma}
The sum of two independent stationary series generated by ARMA($p$, $m$) and ARMA($q$, $n$) is generated by ARMA($x$, $y$), where $x\leq p+q$ and $y\leq \max(p+n,q+m)$.

In shorthand notation, $\textrm{ARMA}(p, m) + \textrm{ARMA}(q, n) = \textrm{ARMA}(p+q, \max(p+n,q+m))$.
\end{lemma}

When two ARMAX processes share the same exogenous input series, the dependency on exogenous input is additive, and the above can be extended to $\textrm{ARMAX}(p, m, r) + \textrm{ARMAX}(q, n, s) = \textrm{ARMAX}(p+q, \max(p+n,q+m), \max(r, s))$.

\paragraph{Jordan canonical form and canonical basis}

Every square real matrix is similar to a complex block diagonal matrix known as its Jordan canonical form (JCF). In the special case for diagonalizable matrices, JCF is the same as the diagonal form. Based on JCF, there exists a canonical basis $\{e_i\}$ consisting only of eigenvectors and generalized eigenvectors of $A$. A vector $v$ is a generalized eigenvector of rank $\mu$ with corresponding eigenvalue $\lambda$ if $(\lambda I - A)^{\mu}v = 0$ and $(\lambda I - A)^{\mu - 1}v \neq 0$.

Relating the canonical basis to the characteristic polynomial, the characteristic polynomial can be completely factored into linear factors $\chi_A(\lambda) = (\lambda - \lambda_1)^{\mu_1}(\lambda - \lambda_2)^{\mu_2}\cdots(\lambda - \lambda_r)^{\mu_r}$ over $\mathbb{C}$. The complex roots $\lambda_1, \cdots, \lambda_r$ are eigenvalues of $A$. For each eigenvalue $\lambda_i$, there exist $\mu_i$ linearly independent generalized eigenvectors $v$ such that $(\lambda_i I - A)^{\mu_i}v = 0$.

\subsection{General model equivalence theorem}

Now we state Theorem \ref{thm:model_eq_detail}, a more detailed version of Theorem \ref{thm:model_eq}. 

\begin{theorem}\label{thm:model_eq_detail}
For any linear dynamical system with parameters $\Theta = (A,B,C,D)$, hidden dimension $n$, inputs $x_t \in \mathbb{R}^k$, and outputs $y_t \in \mathbb{R}^m$, the outputs $y_t$ satisfy
\begin{equation}
\chi_A^{\dagger}(L)y_t = \chi_A^{\dagger}(L)\xi_t + \Gamma(L)x_t,
\end{equation}
where $L$ is the lag operator, $\chi_A^{\dagger}(L) = L^n\chi_A(L^{-1})$ is the reciprocal polynomial of the characteristic polynomial of $A$, and $\Gamma(L)$ is an $m$-by-$k$ matrix of polynomials of degree $n-1$.

This implies that each dimension of $y_t$ can be generated by an ARMAX($n, n, n-1$) model, where the autoregressive parameters are the characteristic polynomial coefficients in reverse order and in negative values.
\end{theorem}

To prove the theorem, we introduce a lemma to analyze the autoregressive behavior of the hidden state projected to a generalized eigenvector direction.

\begin{lemma}\label{lemma:ar_poly_single}
Consider a linear dynamical system with parameters $\Theta = (A,B,C,D)$, hidden states $h_t \in \mathbb{R}^n$, inputs $x_t \in \mathbb{R}^k$, and outputs $y_t \in \mathbb{R}^m$ as defined in (\ref{eq:lds_def}). For any generalized eigenvector $e_i$ of $A^*$ with eigenvector $\lambda$ and rank $\mu$, the lag operator polynomial $(1-\lambda L)^{\mu}$ applied to time series $h_t^{(i)} := \inner{h_t, e_i}$ results in 
$$(1-\lambda L)^{\mu}h_t^{(i)} = \textrm{linear transformation of }x_t, \cdots, x_{t-\mu+1}.$$
\end{lemma}

\begin{proof}
To expand the LHS, first observe that
\begin{align*}
(1-\lambda L)h_t^{(i)} & = (1-\lambda L)\inner{h_t, e_i} \\
& = \inner{h_t^{(i)}, e_i} - \lambda L \inner{h_t^{(i)}, e_i}  \\
& = \inner{Ah_{t-1}+Bx_t, e_i} - \inner{h_{t-1}, \lambda e_i}  \\
& = \inner{h_{t-1}^{(i)}, (A^*-\lambda I) e_i}  + \inner{Bx_t, e_i}.
\end{align*}
We can apply $(1-\lambda L)$ again similarly to obtain
\begin{align*}
%(1-\lambda L)^2 h_t^{(i)} & = (1-\lambda L)\inner{h_{t-1}, (A^*-\lambda I) e_i} + (1-\lambda L) \inner{Bx_t, e_i} \\
(1-\lambda L)^2 h_t^{(i)}  = \inner{h_{t-2}, (A^*-\lambda I)^2 e_i}  \\ + \inner{Bx_{t-1}, (A^*-\lambda I) e_i} + (1-\lambda L)\inner{Bx_t, e_i},
\end{align*}
and in general we can show inductively that
\begin{align*}
(1-\lambda L)^k h_t^{(i)} - \inner{h_{t-k}, (A^*-\lambda I)^ke_i} = \\
 \sum_{j=0}^{k-1} (1-\lambda L)^{k-1-j} L^j \inner{Bx_t, (A^*-\lambda I)^j e_i},
\end{align*}
where the RHS is a linear transformation of $x_t, \cdots, x_{t-k+1}$.

Since $(\lambda I -A^*)^\mu e_i = 0$ by definition of generalized eigenvectors, $\inner{h_{t-\mu}, (A^*-\lambda I)^{\mu}e_i} = 0$, and hence $(1-\lambda L)^{\mu} h_t^{(i)}$ itself is a linear transformation of $x_t, \cdots, x_{t-\mu+1}$.
\end{proof}

\paragraph{Proof for Theorem \ref{thm:model_eq_detail}}

Using Lemma \ref{lemma:ar_poly_single} and the canonical basis, we can prove Theorem \ref{thm:model_eq_detail}.

\begin{proof}
Let $\lambda_1, \cdots, \lambda_r$ be the eigenvalues of $A$ with multiplicity $\mu_1, \cdots, \mu_r$. Since $A$ is a real-valued matrix, its adjoint $A*$ has the same characteristic polynomial and eigenvalues as $A$. There exists a canonical basis $\{e_i\}_{i=1}^n$ for $A^*$, where $e_1, \cdots, e_{\mu_1}$ are generalized eigenvectors with eigenvalue $\lambda_1$, $e_{\mu_1+1}, \cdots, e_{\mu_1+\mu_2}$ are generalized eigenvectors with eigenvalue $\lambda_2$, so on and so forth, and $e_{\mu_1+\cdots+\mu_{r-1}+1}, \cdots, e_{\mu_1+\cdots+\mu_r}$ are generalized eigenvectors with eigenvalue $\lambda_r$.

By Lemma (\ref{lemma:ar_poly_single}), $ (1-\lambda_1 L)^{\mu_1}h_t^{(i)}$ is a linear transformation of $x_t, \cdots, x_{t-\mu_1+1}$ for $i=1,\cdots,\mu_1$; $(1-\lambda_2 L)^{\mu_2}h_t^{(i)}$ is a linear transformation of $x_t, \cdots, x_{t-\mu_2+1}$ for $i=\mu_1+1,\cdots,\mu_1+\mu_2$; so on and so forth; $(1-\lambda_r L)^{\mu_r}h_t^{(i)}$ is a linear transformation of $x_t, \cdots, x_{t-\mu_r+1}$ for $i=\mu_1+\cdots+\mu_{r-1}+1,\cdots,n$.

We then apply lag operator polynomial $\Pi_{j\neq i}(1-\lambda_j L)^{\mu_j}$ to both sides of each equation. The lag polynomial in the LHS becomes 
$(1-\lambda_1 L)^{\mu_1}\cdots(1-\lambda_rL)^{\mu_r} = \chi_A^{\dagger}(L).$
For the RHS, since $\Pi_{j\neq i}(1-\lambda_j L)^{\mu_j}$ is of degree $n - \mu_i$, it lags the RHS by at most $n-\mu_i$ additional steps, and the RHS becomes a linear transformation of $x_t, \cdots, x_{t-n+1}$.

Thus, for each $i$, $\chi_A^{\dagger}(L) h_t^{(i)}$ is a linear transformation of $x_t, \cdots, x_{t-n+1}$.

The outputs of the LDS are defined as $y_t = Ch_t + Dx_t + \xi_t = \sum_{i=1}^n h_t^{(i)} Ce_i + Dx_t + \xi_t$. By linearity, and since $\chi_A^{\dagger}(L)$ is of degree $n$, both $\sum_{i=1}^n h_t^{(i)} Ce_i$ and  $\chi_A^{\dagger}(L) Dx_t$ are linear transformations of $x_t, \cdots, x_{t-n}$. We can write any such linear transformation as $\Gamma(L)x_t$ for some $m$-by-$k$ matrix $\Gamma(L)$ of polynomials of degree $n-1$. Thus, as desired,
\begin{align*}
\chi_A^{\dagger}(L) y_t = & \chi_A^{\dagger}(L) \xi_t + \Gamma(L)x_t.
\end{align*}

Assuming that there are no common factors in $\chi_A^{\dagger}$ and $\Gamma$, $\chi_A^{\dagger}$ is then the lag operator polynomial that represents the autoregressive part of $y_t$. This assumption is the same as saying that $y_t$ cannot be expressed as a lower-order ARMA process.
The reciprocal polynomial has the same coefficients in reverse order as the original polynomial. According to the lag operator polynomial on the LHS, $1 - \varphi_1 L - \varphi_2 L^2 - \cdots - \varphi_n L^n = \chi_A^{\dagger}(L)$, and $L^n - \varphi_1 L^{n-1} - \cdots - \varphi_n = \chi_A(L)$, so the $i$-th order autoregressive parameter $\varphi_i$ is the negative value of the $(n-i)$-th order coefficient in the characteristic polynomial $\chi_A$. 

\end{proof}

\subsection{The hidden input case as a corollary}

The statement about LDS without external inputs in Theorem \ref{thm:model_eq} comes as a corollary to Theorem \ref{thm:model_eq_detail}, with a short proof here.

\begin{proof}

Define $y_t' = Ch_t + Dx_t$ to be the output without noise, i.e. $y_t = y_t' + \xi_t$. By Theorem \ref{thm:model_eq_detail}, $\chi_A^{\dagger}(L)y_t' = \Gamma(L)x_t$. Since we assume the hidden inputs $x_t$ are i.i.d. Gaussians, $y_t'$ is then generated by an ARMA($n, n-1$) process with autoregressive polynomial $\chi_A^{\dagger}(L)$.

The output noise $\xi_t$ itself can be seen as an ARMA($0, 0$) process. By Lemma \ref{lemma:sum_arma}, ARMA($n, n-1$) +  ARMA($0, 0$) = ARMA($n + 0, \max(n + 0, n-1 + 0)$) = ARMA($n, n$). Hence the outputs $y_t$ are generated by an ARMA($n, n$) process as claimed in Theorem \ref{thm:model_eq}. It is easy to see in the proof of Lemma \ref{lemma:sum_arma} that the autoregressive parameters do not change when adding a white noise~\cite{granger1976time}. 
\end{proof}

\section{Proof for eigenvalue approximation theorems}\label{sec:approx_eig}

Here we restate Theorem \ref{thm:approx_eig} and Theorem \ref{thm:explicit_cond_num} together, and prove it in three steps for 1) the general case, 2) the simple eigenvalue case, and 3) the explicit condition number bounds for the simple eigenvalue case.

\begin{theorem}
Suppose $y_t$ are the outputs from an $n$-dimensional latent linear dynamical system with parameters $\Theta=(A,B,C,D)$ and eigenvalues $\lambda_1,\cdots,\lambda_n$. Let $\hat\Phi = (\hat\varphi_1, \cdots, \hat\varphi_n)$ be the estimated autoregressive parameters with error $\norm{\hat\Phi - \Phi} = \epsilon$, and let $r_1, \cdots, r_n$ be the roots of the polynomial $1-\hat\varphi_1z-\cdots-\hat\varphi_nz^n$.

Assuming the LDS is observable, the roots converge to the true eigenvalues with convergence rate $\mathcal{O}(\epsilon^{1/n})$. 
If all eigenvalues of $A$ are simple (i.e. multiplicity 1), then the convergence rate is $\mathcal{O}(\epsilon)$.
If $A$ is symmetric, Lyapunov stable (spectral radius at most 1), and only has simple eigenvalues, then
$$|r_i - \lambda_i| \leq \frac{\sqrt{n2^{n-1}}}{\Pi_{k\neq j}|\lambda_j-\lambda_k|}\epsilon + \mathcal{O}(\epsilon^2).$$
\end{theorem}

\subsection{General (1/n)-exponent bound}

This is a known perturbation bound on polynomial root finding due to Ostrowski~\cite{beauzamy1999roots}.
\begin{lemma}\label{lemma:poly_roots_perturb}
Let $\Phi(z) = z^n + \varphi_1 z^{n-1} + \cdots + \varphi_{n-1}z + \varphi_n$ and $\Psi(z) = z^n + \psi_1 z^{n-1} + \cdots + \psi_{n-1}z + \psi_n$ be two polynomials of degree $n$. If $\norm{\Phi-\Psi}_2 < \epsilon$, then the roots $(r_k)$ of $\Phi$ and roots $(\tilde r_k)$ of $\Psi$ under suitable order satisfy
$$|r_k - \tilde r_k| \leq 4Cp\epsilon^{1/n},$$
where $C = \max_{1, 0\leq k \leq n}\{|\varphi_n|^{1/n}, |\psi_n|^{1/n}\}$.
\end{lemma}

The general $\mathcal{O}(\epsilon^{1/n})$ convergence rate in Theorem \ref{thm:approx_eig} follows directly from Lemma \ref{lemma:poly_roots_perturb} and Theorem \ref{thm:model_eq}.

\subsection{Bound for simple eigenvalues}

The $\frac{1}{n}$-exponent in the above bound might seem not very ideal, but without additional assumptions the $\frac{1}{n}$-exponent is tight. As an example, the polynomial $x^2 - \epsilon$ has roots $x \pm \sqrt{\epsilon}$. This is a general phenomenon that a root with multiplicity $m$ could split into $m$ roots at rate $O(\epsilon^m)$, and is related to the \textit{regular splitting property}~\cite{hryniv1999perturbation, lancaster2003perturbation} in matrix eigenvalue perturbation theory. 

Under the additional assumption that all the eigenvalues are simple (no multiplicity), we can prove a better bound using the following idea with companion matrix: Small perturbation in autoregressive parameters results in small perturbation in companion matrix, and small perturbation in companion matrix results in small perturbation in eigenvalues.

\paragraph{Matrix eigenvalue perturbation theory}

The perturbation bound on eigenvalues is a well-studied problem~\cite{greenbaum2019first}. The \textit{regular splitting property} states that, for an eigenvalue $\lambda_0$ with partial multiplicities $m_1, \cdots, m_k$, an $O(\epsilon)$ perturbation to the matrix could split the eigenvalue into $M = m_1 + \cdots + m_k$ distinct eigenvalues $\lambda_{ij}(\epsilon)$ for $i=1,\cdots,k$ and $j=1,\cdots,m_i$, and each eigenvalue $\lambda_{ij}(\epsilon)$ is moved from the original position by $O(\epsilon^{1/m_i})$.

For semi-simple eigenvalues, geometric multiplicity equals algebraic multiplicity. Since geometric multiplicity is the number of partial multiplicities while algebraic multiplicity is the sum of partial multiplicities, for semi-simple eigenvalues all partial multiplicities $m_i = 1$. Therefore, the regular splitting property corresponds to the asymptotic relation in equation \ref{eq:eig_split}.
It is known that regular splitting holds for any semi-simple eigenvalue even for non-Hermitian matrices.

\begin{lemma}[Theorem 6 in~\cite{lancaster2003perturbation}]\label{lemma:semisimple_perturb}
Let $L(\lambda, \epsilon)$ be an analytic matrix function with semi-simple eigenvalue $\lambda_0$ at $\epsilon=0$ of multiplicity $M$. Then there are exactly $M$ eigenvalues $\lambda_i(\epsilon)$ of $L(\lambda, \epsilon)$ for which $\lambda_i(\epsilon) \rightarrow \lambda_0$ as $\epsilon \rightarrow 0$, and for these eigenvalues
\begin{equation}\label{eq:eig_split}
\lambda_{i}(\epsilon) = \lambda_0 + \lambda_{i}'\epsilon + o(\epsilon).
\end{equation}
\end{lemma}

\paragraph{Companion Matrix}

Matrix perturbation theory tell us how perturbations on matrices change eigenvalues, while we are interested in how perturbations on polynomial coefficients change roots. To apply matrix perturbation theory on polynomials, we introduce the \textit{companion matrix}, also known as the \textit{controllable canonical form} in control theory.

\begin{definition}

For a monic polynomial $\Phi(u) = z^n + \varphi_1 z^{n-1} + \cdots + \varphi_{n-1}z + \varphi_n$, the companion matrix of the polynomial is the square matrix
    \vspace{5em}

$$C(\Phi)={\begin{bmatrix}0&0&\dots &0&-\varphi_{n}\\1&0&\dots &0&-\varphi_{n-1}\\0&1&\dots &0&-\varphi_{n-2}\\\vdots &\vdots &\ddots &\vdots &\vdots \\0&0&\dots &1&-\varphi_{1}\end{bmatrix}}.$$

The matrix $C(\Phi)$ is the companion in the sense that its characteristic polynomial is equal to $\Phi$.
\end{definition}

%When $\Phi$ has distinct roots $\lambda_1, \cdots, \lambda_n$, the companion matrix $C(\Phi)$ is diagonalizable as
%\begin{equation}
%C(\Phi) = V^{-1}\text{diag}(\lambda_1, \cdots, \lambda_n)V,
%\end{equation}
%where $V$ is the Vandermonde matrix corresponding to the $\lambda$'s. Note that $V$ is not an orthogonal matrix.

In relation to a pure autoregressive AR($p$) model, the companion matrix corresponds to the transition matrix in the linear dynamical system when we encode the values form the past $p$ lags as a $p$-dimensional state
$$h_t = {\begin{bmatrix}y_{t-p+1} & \cdots & y_{t-1} & y_t\end{bmatrix}}^T.$$ If $y_{t} = \varphi_1 y_{t-1} + \cdots + \varphi_p y_{t-p}$, then $h_{t} =$
\begin{equation}\label{eq:sys_from_companion}
\begin{split}
&{\begin{bmatrix}y_{t-p+1}\\y_{t-p+2}\\ \cdots \\ y_{t-1} \\ y_{t}\end{bmatrix}}
= {\begin{bmatrix}0&1&0&\dots &0\\0&0&1&\dots &0\\ \vdots&\vdots&\vdots&\ddots&\vdots \\0&0&0&\dots &1\\ \varphi_p&\varphi_{p-1}&\varphi_{p-2}&\dots &\varphi_{1}\end{bmatrix}}
  {\begin{bmatrix}y_{t-p}\\y_{t-p+1}\\ \cdots \\ y_{t-2} \\ y_{t-1}\end{bmatrix}} \\
&= C(-\Phi)^T h_{t-1}.
\end{split}
\end{equation}

\paragraph{Proof of Theorem \ref{thm:approx_eig} for simple eigenvalues}

\begin{proof}
Let $y_t$ be the outputs of a linear dynamical system $S$ with only simple eigenvalues, and let $\Phi = (\varphi_1, \cdots, \varphi_n)$ be the ARMAX autoregressive parameters for $y_t$. Let $C(\Phi)$ be the companion matrix of the polynomial $z^n-\varphi_1 z^{n-1} - \varphi_2 z^{n-2} - \cdots - \varphi_n$. The companion matrix is the transition matrix of the LDS described in equation \ref{eq:sys_from_companion}. Since this LDS the same autoregressive parameters and hidden state dimension as the original LDS, by Corollary \ref{cor:char_poly_determines_ar_params} the companion matrix has the same characteristic polynomial as the original LDS, and thus also has simple (and hence also semi-simple) eigenvalues. The $O(\epsilon)$ convergence rate then follows from Lemma \ref{lemma:semisimple_perturb} and Theorem \ref{thm:armax_convergence}, as the error on ARMAX parameter estimation can be seen as perturbation on the companion matrix.
\end{proof}

\paragraph{A note on the companion matrix}One might hope that we could have a more generalized result using Lemma \ref{lemma:semisimple_perturb} for all systems with semi-simple eigenvalues instead of restricting to matrices with simple eigenvalues. Unfortunately, even if the original linear dynamical system has only semi-simple eigenvalues, in general the companion matrix is not semi-simple unless the original linear dynamical system is simple. This is because the companion matrix always has its minimal polynomial equal to its characteristic polynomial, and hence has geometric multiplicity 1 for all eigenvalues. This also points to the fact that even though the companion matrix has the form of the controllable canonical form, in general it is not necessarily similar to the transition matrix in the original LDS.

\subsection{Explicit bound for condition number}

In this subsection, we write out explicitly the condition number for simple eigenvalues in the asymptotic relation $\lambda(\epsilon) = \lambda_0 + \kappa\epsilon + o(\epsilon)$, to show how it varies according to the spectrum. Here we use the notation $\kappa(C, \lambda)$ to note the condition number for eigenvalue $\lambda$ in companion matrix $C$.

\begin{lemma}\label{lemma:explicit_cond_num}
For a companion matrix $C$ with simple eigenvalues $\lambda_1, \cdots, \lambda_n$, the eigenvalues $\lambda_1', \cdots, \lambda_n'$ of the perturbed matrix by $C + \delta C$ satisfy
\begin{equation}
|\lambda_j - \lambda_j'| \leq \kappa(C, \lambda_j)\norm{\delta C}_2 + o(\norm{\delta C}_2^2),
\end{equation}
and the condition number $\kappa(C, \lambda_j)$ is bounded by
\begin{equation}
\begin{split}
\frac{1}{\prod_{k \neq j} |\lambda_j - \lambda_k|} \leq
\kappa(C, \lambda_j) \leq \\
\frac{\sqrt{n}}{\prod_{k \neq j} |\lambda_j - \lambda_k|} \left(\max(1, |\lambda_j|)\right)^{n-1} (1 + \rho(C)^2)^{\frac{n-1}{2}},
\end{split}
\end{equation}
where $\rho(C)$ is the spectral radius, i.e. largest absolute value of its eigenvalues.

In particular, when $\rho(C) \leq 1$, i.e. when the matrix is Lyapunov stable,
\begin{equation}
|\lambda_j - \lambda_j'| \leq \frac{\sqrt{n} (\sqrt{2})^{n-1}}{\prod_{k \neq j} |\lambda_j - \lambda_k|}\norm{\delta C}_2 + o(\norm{\delta C}_2^2).
\end{equation}
\end{lemma}

\begin{proof}

For each simple eigenvalue $\lambda$ of the companion matrix $C$ with column eigenvector $v$ and row eigenvector $w^*$, the condition number of the eigenvalue is
\begin{equation}
\kappa(C, \lambda) = \frac{\norm{w}_2\norm{v}_2}{|w^* v|}.
\end{equation}

This is derived from differentiating the eigenvalue equation $Cv = v\lambda$, and multiplying the differentiated equation by $w*$, which results in
$$ w^* (\delta C)v + w^* C(\delta v) = \lambda w^* (\delta v) + w^* v(\delta \lambda).$$%Since $w^*C = \lambda w^*$, we then have
$$\delta\lambda = \frac{w^*(\delta C) v}{w^*v}.$$
Therefore,
\begin{equation}
|\delta \lambda| \leq \frac{\norm{w}_2\norm{v}_2}{|w^*v|} \norm{\delta C}_2 = \kappa(C, \lambda) \norm{\delta C}_2.
\end{equation}

The companion matrix can be diagonalized as $C = V^{-1}\text{diag}(\lambda_1, \cdots, \lambda_n)V$, the rows of the Vandermonde matrix $V$ are the row eigenvectors of $C$, while the columns of $V^{-1}$ are the column eigenvectors of $C$. Since the the $j$-th row $V_{j, *}$ and the $j$-th column $V^{-1}_{*, j}$ have inner product 1 by definition of matrix inverse, the condition number is given by
\begin{equation}\label{condition_number}
\kappa(C, \lambda_j) = \norm{V_{j, *}}_2 \ \norm{V^{-1}_{*, j}}_2.
\end{equation}

\paragraph{Formula for inverse of Vandermonde matrix}
The Vandermonde matrix is defined as
\begin{equation}
V = {\begin{bmatrix}
        1 & \lambda_1 & \lambda_1^2 & \cdots & \lambda_1^{p-1} \\
        1 & \lambda_2 & \lambda_2^2 & \cdots & \lambda_2^{p-1} \\
        \vdots & \vdots & \vdots & \cdots & \vdots \\
        1 & \lambda_p & \lambda_p^2 & \cdots & \lambda_p^{p-1} \\
    \end{bmatrix}}.
\end{equation}
The inverse of the Vandermonde matrix $V$ is given by~\cite{el2003explicit} using elementary symmetric polynomial.
\begin{equation}\label{vandermonde_inv}
(V^{-1})_{i,j} = \frac{(-1)^{i+j}  S_{p-i, j}}{\prod_{k < j} (\lambda_j - \lambda_k) \prod_{k > j} (\lambda_k - \lambda_j)},
\end{equation}
where $S_{p-i,j} = S_{p-i}(\lambda_1, \cdots, \lambda_{j-1}, \lambda_{j+1}, \cdots, \lambda_p).$

Pulling out the common denominator, the $j$-th column vector of $V^{-1}$ is
$$\frac{(-1)^{j}}{\prod_{k < j} (\lambda_j - \lambda_k) \prod_{k > j} (\lambda_k - \lambda_j)}
{\begin{bmatrix} (-1)S_{p-1} \\ (-1)^2S_{p-2} \\ \vdots \\ (-1)^{p-1}S_{1} \\ (-1)^{p} \end{bmatrix}},$$ where the elementary symmetric polynomials are over variables $\lambda_1, \cdots, \lambda_{j-1}, \lambda_{j+1}, \cdots, \lambda_p$.

For example, if $p = 4$, then the 3rd column (up to scaling) would be
$$ \frac{-1}{(\lambda_3-\lambda_1)(\lambda_3-\lambda_2)(\lambda_4-\lambda_3)}
{\begin{bmatrix} -\lambda_1\lambda_2\lambda_4 \\ \lambda_1\lambda_2 +\lambda_1\lambda_4 +\lambda_2\lambda_4 \\ -\lambda_1-\lambda_2-\lambda_4 \\ 1 \end{bmatrix}}.$$

\paragraph{Bounding the condition number}

As discussed before, the condition number for eigenvalue $\lambda_j$ is
\begin{equation*}
\kappa(C, \lambda_j) = \norm{V_{j, *}}_2 \ \norm{V^{-1}_{*, j}}_2.
\end{equation*}
where $V_{j, *}$ is the $j$-th row of the Vandermonde matrix $V$ and $V^{-1}_{*, j}$ is the $j$-th column of $V^{-1}$.

By definition $V_{j, *} = {\begin{bmatrix}1 & \lambda_j & \lambda_j^2 & \cdots & \lambda_j^{p-1} \end{bmatrix}}$, so $$\norm{V_{j, *}}_2 = \left(\sum_{i=0}^{p-1} \lambda_j^{2i}\right)^{1/2}.$$

Using the above explicit expression for $V^{-1}$, $\norm{V^{-1}_{*, j}}_2 = $
$$
\frac{1}{\prod_{k \neq j} |\lambda_j - \lambda_k|}
\left(\sum_{i=0}^{p-1}S_i^2(\lambda_1, \cdots, \lambda_{j-1}, \lambda_{j+1}, \cdots, \lambda_p)\right)^{1/2}.
$$.

Therefore, 
\begin{equation}\label{condition_number_explicit}
\begin{split}
\kappa(C, \lambda_j) = \; & \frac{1}{\prod_{k \neq j} |\lambda_j - \lambda_k|} \left(\sum_{i=0}^{p-1} \lambda_j^{2i}\right)^{1/2} \\
& \left(\sum_{i=0}^{p-1}S_i^2(\lambda_1, \cdots, \lambda_{j-1}, \lambda_{j+1}, \cdots, \lambda_p)\right)^{1/2}.
\end{split}
\end{equation}

Note that both parts under $(\cdots)^{1/2}$ are greater than or equal to $1$, so we can bound it below by
$$ \kappa(C, \lambda_j)  \geq \frac{1}{\prod_{k \neq j} |\lambda_j - \lambda_k|}.$$

We could also bound the two parts above. The first part can be bounded by
\begin{equation}\label{eq:cond_num_upper_part_1}
\left(\sum_{i=0}^{p-1} \lambda_j^{2i}\right)^{1/2} \leq \sqrt{p} \max(1, |\lambda_j|)^{(p-1)}.
\end{equation}
While for the second part, since
$$ |S_i(\lambda_1, \cdots, \lambda_{j-1}, \lambda_{j+1}, \cdots, \lambda_p)| \leq
\binom{p-1}{i}|\lambda|_{\max}^i,$$
we have that
\begin{equation}\label{eq:cond_num_upper_part_2}
\begin{split}
\sum_{i=0}^{p-1}S_i^2(\lambda_1, \cdots, \lambda_{j-1}, \lambda_{j+1}, \cdots, \lambda_p)
\\ \leq \sum_{i=0}^{p-1} \binom{p-1}{i}|\lambda|_{\max}^{2i} = (1 + |\lambda|_{\max}^2)^{p-1}.
\end{split}
\end{equation}
Combining equation \ref{eq:cond_num_upper_part_1} and \ref{eq:cond_num_upper_part_2} for the upper bound, and putting it together with the lower bound,
%$$ \kappa(C, \lambda_j) \leq \frac{\sqrt{p}}{\prod_{k \neq j} |\lambda_j - \lambda_k|} \left(\max(1, |\lambda_j|)\right)^{p-1} (1 + |\lambda|_{\max}^2)^{\frac{p-1}{2}}.$$

\begin{equation}
\begin{split}
\frac{1}{\prod_{k \neq j} |\lambda_j - \lambda_k|} \leq
\kappa(C, \lambda_j) \leq \\
\frac{\sqrt{p}}{\prod_{k \neq j} |\lambda_j - \lambda_k|} \left(\max(1, |\lambda_j|)\right)^{p-1} (1 + \rho(C)^2)^{\frac{p-1}{2}},
\end{split}
\end{equation}
as desired.
\end{proof}

Theorem \ref{thm:explicit_cond_num} follows from Lemma \ref{lemma:explicit_cond_num}, because the estimation error on the autoregressive parameters can be seen as the perturbation on the companion matrix, and the companion matrix has the same eigenvalues as the original LDS.

\section{Iterated regression for ARMAX}\label{sec:appendix_armax_estimation}

\paragraph{Algorithm}

We generalize Algorithm \ref{algo:iter_reg_1d} to accommodate for exogenous inputs. Since the exogenous inputs are explicitly observed, including exogenous inputs in the regression does not change the consistent property of the estimator. 

Theorem \ref{thm:model_eq_detail} shows that different output channels from the same LDS have the same autoregressive parameters in ARMAX models. Therefore, we could leverage multidimensional outputs by estimating the autoregressive parameters in each channel separately and average them. 

\begin{algorithm}
\SetAlgoLined
Input: A time series $\{y_t\}_{t=1}^{T}$ where $y_t \in \mathbb{R}^m$, exogenous input series $\{x_t\}_{t=1}^T$ where $x_t \in \mathbb{R}^k$, and guessed hidden state dimension $n$. \\
\For{$d=1,\cdots,m$} {
Let $y_t^{(d)}$ be the projection of $y_t$ to the $d$-th dimension\;
Initialize error term estimates $\hat \epsilon_t = \vec{0} \in \mathbb{R}^m$ for $t = 1, \cdots, T$\;
\For{$i=0,\cdots,n$} {
Perform $\ell_2$-regularized least squares regression on $y_t$ against lagged terms of $y_t$, $x_t$, and $\hat\epsilon_t$ to solve for coefficients $\hat\varphi_j \in \mathbb{R}$, $\hat\theta_j \in \mathbb{R}$, and $\hat\gamma_j \in \mathbb{R}^{k}$ in the linear equation
$y_t^{(d)} = c + \sum_{j=1}^n \hat\varphi_j y_{t-j}^{(d)} + \sum_{j=1}^{n-1} \hat\gamma_j x_{t-j} + \sum_{j=1}^i \hat\theta_j \hat\epsilon_{t-j}$, with $\ell_2$-regularization only on $\hat\theta_j$\;
Update $\hat\epsilon_t$ to be the residuals from the most recent regression\;
}
Record $\hat\Phi^{(d)} = (\hat\varphi_1, \cdots, \hat\varphi_n)$;
}
Return the average estimate $\hat\Phi = \frac{1}{d}(\hat\Phi^{(1)}+\cdots+\hat\Phi^{(m)})$.
 \caption{Regularized iterated regression for AR parameter estimation in ARMAX}
 \label{algo:iter_reg_general}
\end{algorithm}

Again as before the $i$-th iteration of the regression only uses error terms from the past $i$ lags. In other words, the initial iteration is an ARMAX($n,0,n-1$) regression, the first iteration is an ARMAX($n,1,n-1$) regression, and so forth.

\paragraph{Time complexity}
The iterated regression in each dimension involves $n+1$ steps of least squares regression each on at most $n(k+2)$ variables. Therefore, the total time complexity of Algorithm \ref{algo:iter_reg_general} is $O(nm((nk)^2T + (nk)^3)) = O(mn^3k^2T + mn^4k^3)$, where $T$ is the sequence length, $n$ the hidden state dimension, $m$ the output dimension, and $k$ the input dimension.

\section{Additional simulation details}

\subsection{Synthetic data generation}\label{sec:data_generation}

%We generate clusters of linear dynamical systems by the following procedure. 

First, we generate $K$ cluster centers by generating LDSs with random matrices $A, B, C$ of standard i.i.d. Gaussians. We assume that the output $y_t$ only depends on the hidden state $h_t$ but not the input $x_t$, i.e. the matrix $D$ is zero. When generating the random LDSs, we require that the spectral radius $\rho(A) \leq 1$, i.e. all eigenvalues of $A$ have absolute values at most 1, and regenerate a new random matrix if the spectral radius is above 1. Our method also applies to the case of arbitrary spectral radius, this requirement is for the purpose of preventing numeric overflow in generated sequence. We also require that the $\ell_2$ distance $d(\Theta_1, \Theta_2) = \norm{\lambda(A_1) - \lambda(A_2)}_2$ between cluster centers are at least 0.2 apart.

Then, we generate 100 LDSs by randomly assigning them to the clusters. To obtain a LDS with assigned cluster center $\Theta = (A_c, B_c, C_c)$, we generate $A'$ by adding a i.i.d. Gaussians to each entry of $A_c$, while $B'$ and $C'$ are new random matrices of i.i.d. standard Gaussians. The standard deviation of the i.i.d. Gaussians for $A' - A_c$ is chosen such that the average distance to cluster centers is less than half of the inter-cluster distance between centers.

For each LDS, we generate a sequence by drawing hidden inputs $x_t \sim N(0, 1)$ and put noise $\xi_t \sim N(0, 0.01^2)$ on the outputs.

\subsection{Empirical correlation between AR distance and LDS distance.}
\begin{figure}[b]
    \centering
    \includegraphics[width=\linewidth]{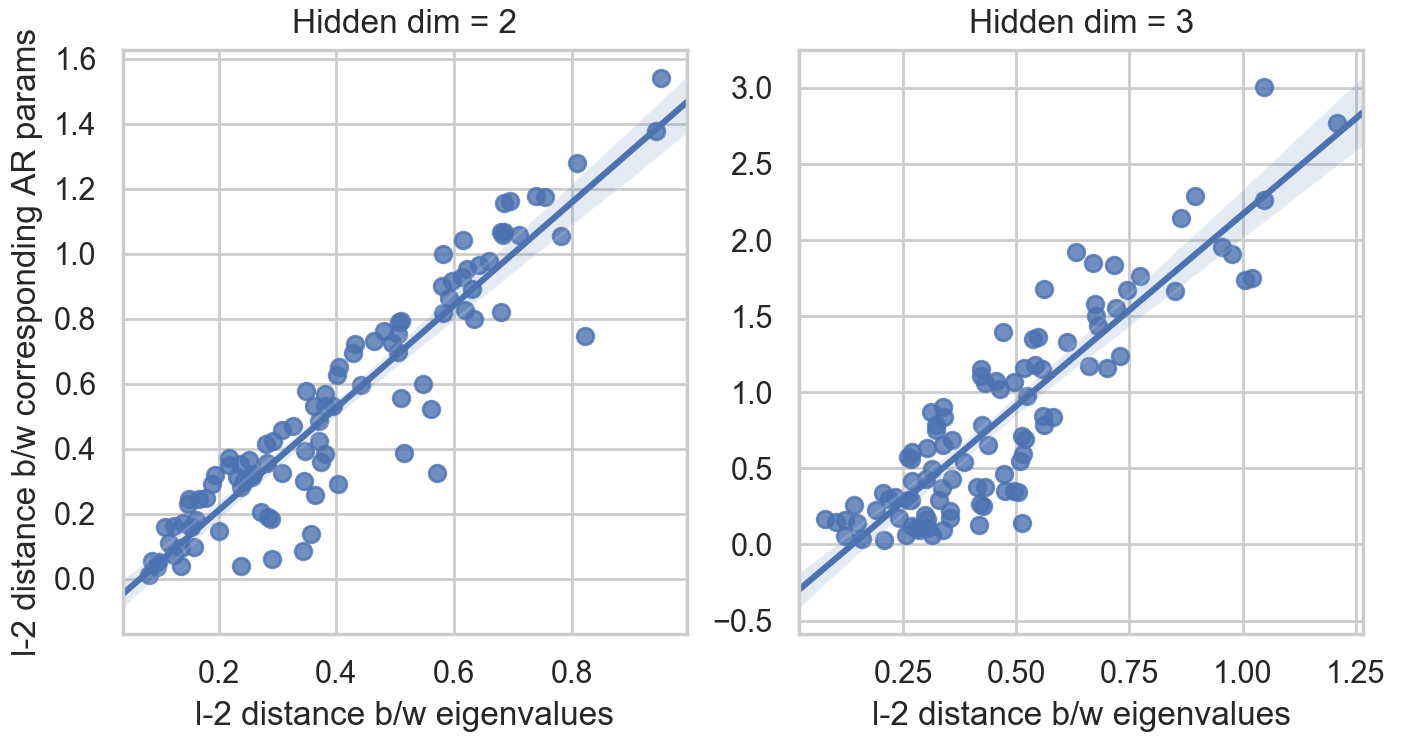}
    \caption{The eigenvalue $\ell_2$ distance and the autoregressive parameter $\ell_2$ distance for 100 random linear dynamical systems with eigenvalues drawn uniformly randomly from $[-1, 1]$. The two distance measures are highly correlated.}
    \label{fig:ar_dist_to_eig_dist}
\end{figure}

Theorem~\ref{thm:approx_eig} shows that LDSs with similar AR parameters also have similar eigenvalues.
The converse of Theorem~\ref{thm:approx_eig} is also true: dynamical systems with small eigenvalue distance have small autoregressive parameter distance, which follows from perturbation bounds for characteristic polynomials~\cite{ipsen2008perturbation}.
Figure~\ref{fig:ar_dist_to_eig_dist} shows simulation results where the AR parameter distance and the LDS eigenvalue distance are highly correlated.

\end{document}